\pdfoutput=1

\documentclass[11pt]{article}

\usepackage[]{acl}

\usepackage{times}
\usepackage{latexsym}

\usepackage[T1]{fontenc}

\usepackage[utf8]{inputenc}

\usepackage{microtype}

\usepackage{inconsolata}

\usepackage{xspace,mfirstuc,tabulary}

\usepackage{graphicx}
\usepackage{breqn}
\usepackage{amssymb}

\usepackage{inconsolata}

\usepackage{bibentry}
\usepackage{dialogue}
\usepackage{times}
\usepackage{latexsym}
\usepackage{mathrsfs}
\usepackage{amssymb}
\usepackage{amsmath}
\usepackage{url}
\usepackage{pifont}
\usepackage{graphicx}
\usepackage{xcolor}
\usepackage{braket}
\usepackage{bbold}
\usepackage[utf8]{inputenc}
\usepackage[toc,page]{appendix}
\usepackage{cleveref}
\usepackage{booktabs,graphicx,makecell,siunitx,eqparbox}
\usepackage[utf8]{inputenc}
\usepackage{color,soul}
\usepackage{arydshln}
\usepackage{enumitem}
\usepackage{colortbl}
\usepackage[most]{tcolorbox}
\usepackage{caption}
\usepackage{multirow}
\usepackage{array}
\usepackage{algorithm}
\usepackage{algpseudocode}
\usepackage{placeins} 
\usepackage{float}

\usepackage{stfloats}

\definecolor{Red}{RGB}{224,102,102}
\definecolor{Green}{RGB}{147,196,125}
\definecolor{Blue}{RGB}{102,178,255}

\usepackage{a4wide}
\newtheorem{definition}{Definition}

\title{Less for More: Enhanced Feedback-aligned Mixed LLMs for Molecule Caption Generation and Fine-Grained NLI Evaluation}

\author{
 \textbf{Dimitris Gkoumas\textsuperscript{1}},
 \textbf{Maria Liakata\textsuperscript{1,2}}
\\
\\
 \textsuperscript{1}Queen Mary University of London, London, UK,\\
 \textsuperscript{2}The Alan Turing Institute, London, UK
\\
  \{d.gkoumas, m.liakata\}@qmul.ac.uk\\
}

\begin{document}
\maketitle
\begin{abstract}
Scientific language models drive research innovation but require extensive fine-tuning on large datasets. This work enhances such models by improving their inference and evaluation capabilities with minimal or no additional training. Focusing on molecule caption generation, we explore post-training synergies between alignment fine-tuning and model merging in a cross-modal setup. We reveal intriguing insights into the behaviour and suitability of such methods while significantly surpassing state-of-the-art models. Moreover, we propose a novel atomic-level evaluation method leveraging off-the-shelf Natural Language Inference (NLI) models for use in the unseen chemical domain. Our experiments demonstrate that our evaluation operates at the right level of granularity, effectively handling multiple content units and subsentence reasoning, while widely adopted NLI methods consistently misalign with assessment criteria.

\end{abstract}

\section{Introduction}
\label{sec:introduction}
AI in Chemistry is essential for developing scalable and cost-effective scientific solutions, such as pioneering drugs~\cite{ferguson2018kinase}, advanced materials~\cite{kippelen2009organic}, and improved chemical processes~\cite{zhong2023reaction}. The vast search spaces in which these solutions reside make chemical language models crucial for accelerating scientific discovery~\cite{ai4science2023impact,zhang2307artificial}. Recent trends have led to the use of multimodal models to learn molecular and linguistic representations, either in separate but coordinated spaces~\cite{edwards2021text2mol,edwards2022translation,liu2023multi}, in a common space~\cite{liu2023molxpt}, or through dual approaches~\cite{luo2023molfm,christofidellis2023unifying}. These models often rely heavily on extensive supervised fine-tuning. However, merely increasing model size and data does not guarantee improvement~\cite{tirumala2022memorization,xu2023paradigm}. Thus, we propose focusing on novel training methods.

\begin{figure}[!ht] 
  \centering
  \includegraphics[width=\columnwidth]{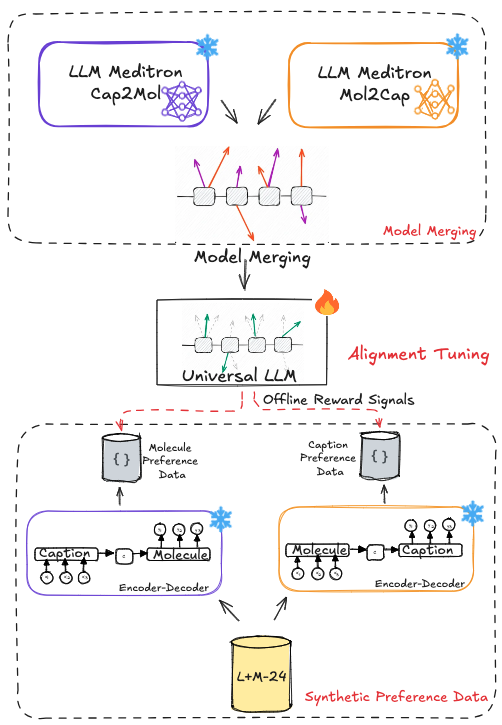} 
  \caption{Overview of our proposed post-training approach to address key limitations in chemical LLMs. Top: Merging per-task pretrained models to create a universal model (refer to $\S~\ref{sec:model_merging}$). Bottom: Generating synthetic preference data using pretrained per-task encoder–decoders (refer to $\S~\ref{sec:setup}$) for alignment tuning . } 
  \label{fig:framework}
\end{figure}
Here we enhance molecule language models using minimal post-training by leveraging synergies between alignment fine-tuning~\cite{ouyang2022training} and model merging~\cite{yang2024model} in a crossmodal setup. Specifically, we focus on molecule-language translation, using as little as 10\% of the training data~\cite{edwards2024l+}. Fig.~\ref{fig:framework} illustrates our comprehensive post-training solution.

Model merging, a technique for fusing models fine-tuned on different tasks, builds a versatile model without needing the original training data or expensive computation. This method has been quickly adopted in foundation language models~\cite{yang2024model}. We extend this concept to a crossmodal setting by merging per-task pretrained molecule language models (see Fig.~\ref{fig:framework}), deploying both weight- and subspace-based techniques to obtain universal models (\S~\ref{sec:model_merging}).

For fine-tuning alignment, we focus on Reinforcement Learning from Human Feedback (RLHF)~\cite{stiennon2020learning} to align the universal models. Although alignment has typically been used to calibrate LLM behaviour~\cite{askell2021general}, we hypothesise that it can also accelerate learning in crossmodal spaces by rewarding preferred over dispreferred outputs, thus improving inference with minimal training data. We focus on optimisation algorithms using closed-form losses on offline preferences, such as Direct Preference Optimisation (DPO)~\cite{rafailov2024direct}, Contrastive Preference Optimisation (CPO)~\cite{xu2024contrastive}, and Kahneman-Tversky Optimisation (KTO)~\cite{ethayarajh2024kto}. We incorporate golden data as human preferences
and dispreferred synthetic outputs generated by proprietary models into the reward signal (see Fig.~\ref{fig:framework}).

We evaluate our models on out-of-distribution data using established statistical-based metrics~\cite{setsbenchmarking,edwards2022translation}. Additionally, we use Natural Language Inference (NLI) models to assess generated text within the chemical domain. However, off-the-shelf NLI models are suboptimal because: a) they are trained on short texts~\cite{williams2018broad}, while generated outputs may mix overlapping content units~\cite{nenkova2007pyramid}; b) they struggle with unseen domains~\cite{mcintosh2024inadequacies}; and c) they lack subsentence inference, limiting their handling of reordered content (see Fig.~\ref{fig:evaluation}). Thus, we propose a novel atomic-level cross-NLI approach that addresses these issues. By decomposing reference and generated texts into atomic premises and hypotheses using an LLM, we calculate probability distributions of contradiction and entailment via an NLI model and finally apply row-wise operations to obtain novel hallucination and coverage metrics ($\S$\ref{sec:atomic_nli}). 

Our findings and contributions are as follows:
\begin{itemize}[noitemsep,topsep=0pt,parsep=0pt,partopsep=0pt,leftmargin=*] 
    \item  \textbf{Extensive training doesn't guarantee better models.} Models trained on large benchmark datasets exhibit memorisation effects, with performance dropping by 50\% to 100\% on out-of-distribution data ($\S~\ref{sec:alignemt_results}$).
    \item  \textbf{Alignment fine-tuning is not a panacea.} Our experiments reveal that not all fine-tuning approaches applicable to heavily trained models are effective with minimal training ($\S~\ref{sec:alignemt_results}$). 
    \item  \textbf{Effective alignment methods balance structured learning and generalisation.} Of the alignment fine-tuning methods, only CPO managed both crossmodal agnostic and minimal training effectively ($\S~\ref{sec:alignemt_results}$).
    \item  \textbf{Model merging addresses inherent limitations in alignment fine-tuning}. It improves performance with minimal training, reduces dependence on human-labelled data, and provides a scalable, cost-effective alignment method for LLMs. ($\S~\ref{sec:merging}$).
    \item \textbf{Our novel atomic-level cross-NLI evaluation reveals intriguing insights about performance interpretability and effectively handles multiple content units in text.} By contrast, widely adopted NLI methods consistently misalign with assessment criteria ($\S~\ref{sec:atomic_results}$). 
\end{itemize}

\section{Related Work}

\subsection{LLMs for Chemistry}

Existing approaches for LLMs in the chemical domain typically rely on costly pretraining with large unimodal datasets for reaction prediction and retrosynthesis~\cite{schwaller2019molecular, vaucher2020automated}, or task-specific fine-tuning for language-molecule learning~\cite{edwards2021text2mol, edwards2022translation, edwards2024l+} and molecule editing~\cite{liu2023multi, fang2023domain}. Other methods focus on multitask learning, which requires resource-intensive pretraining and large multitask datasets~\cite{lu2022unified, ross2022large, christofidellis2023unifying, zhang2024chemllm}. In contrast, we investigate synergies between fine-tuning alignment~\cite{gkoumas2024almol} and model merging to enhance molecule language models with minimal training.

\subsection{Model Merging}
Existing model merging techniques can be broadly categorised into weight-based, subspace-based, and routing-based approaches. Weight-based methods often use optimisation algorithms~\cite{yang2023adamerging, akiba2024evolutionary} or geometric interpolations~\cite{zhou2024metagpt, goddard2024arcee} to determine optimal task vector coefficients. Subspace-based methods involve pruning~\cite{yadav2023resolving, yu2024language} or masking~\cite{wang2024localizing} to remove insignificant parameters, reducing task interference. Routing-based methods combine models adaptively during inference based on specific input~\cite{muqeeth2023soft, tang2024merging}. We experiment with weight- and subspace-based merging in a crossmodal context.

\subsection{Aligning LLMs}
LLM alignment methods can be divided into test-time and fine-tuning approaches. Test-time alignment techniques, such as prompt engineering and guided decoding~\cite{khanov2024args, huang2024deal}, adjust LLMs without changing their weights, but depend on the original model's performance. Fine-tuning methods, like RLHF~\cite{stiennon2020learning, ouyang2022training}, are effective but complex, requiring model retraining and continuous sampling. DPO~\cite{rafailov2024direct} simplifies RLHF by directly optimizing PPO's objective, while CPO~\cite{xu2024contrastive} improves efficiency by using a uniform reference model. Other methods leverage SFT for optimizing RLHF management and parameter tuning~\cite{ethayarajh2024kto, meng2024simpo}. Here, we explore alignment fine-tuning in a crossmodal setup.

\subsection{NLI-based Evaluation}
NLI models determine the relationship between a \textit{premise} and a \textit{hypothesis}. Existing approaches either identify a sentence in the reference text as the premise (sentence-level NLI)\cite{nie2019adversarial, laban2022summac}, or use the entire reference as the premise~\cite{dziri2022evaluating, honovich2022true}, which can be inefficient for long texts~\cite{schuster2022stretching}. Context-level NLI addresses this by retrieving relevant sentences to create a short context~\cite{nie2019combining, schuster2022stretching, kamoi2023wice}, but lacks sufficient granularity~\cite{nenkova2007pyramid}. We propose a novel atomic-level NLI evaluation for the chemical domain to address these limitations.

\section{Methodology}

\subsection{Task Definition}
\label{sec:task}
Let $(x,y)$ represent a pair of source and target sequences mapped to the $\mathrm{X}$ and $\mathrm{Y}$ spaces, respectively. We cast molecule caption generation (MoCG) as a crossmodal alignment task that operates on offline preference data $\mathcal{D}=\{x^{(i)},y_w^{(i)},y_l^{(i)}\}_{i=1}^N$, where $x$ is the input, and $y_w$ and $y_l$ are the preferred and dispreferred outputs, respectively, with $N$ being the total number of pairs in $\mathcal{D}$. The goal is to learn an optimal function $f: X \leftrightarrow Y$ via a model $\pi_{\theta}$ parameterised by $\theta$. We coordinate the molecule and caption generation tasks via instruction modelling~\footnote{Instructions can be found in Appx.~\ref{appx:instructions}.}.

\subsection{Aligned Mixed Molecule Language Models}
This section elaborates on how we obtain aligned universal molecule language models.

\subsubsection{Universal Models via Model Merging}
\label{sec:model_merging}
\begin{figure}[!ht] 
  \centering
  \includegraphics[width=\columnwidth]{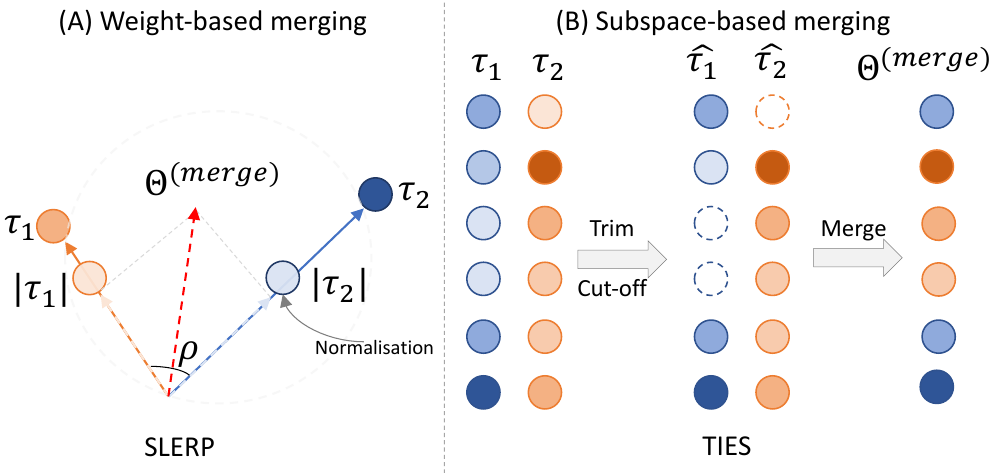} 
  \caption{Model merging techniques for obtaining universal models. (A) Weight-based merging via spherical interpolation. (B) Subspace-based merging by pruning and merging parameter magnitudes. $\tau_1$ and $\tau_2$ are task vectors obtained from pretrained molecule and caption generation models, respectively. }
  \label{fig:merging}
\end{figure}

Let $\tau_1$ and $\tau_2$ represent task vectors~\footnote{A task vector $\tau$ represents the model's parameters $\Theta^{(t)}$ fine-tuned for task $t$~\cite{ilharco2022editing}.} from pretrained molecule and caption generation models. Our goal is to obtain a multitasking cross-modal model $\Theta^{(merge)}$ without accessing training data by exploring weight-based and subspace-based merging techniques. Fig.~\ref{fig:merging} illustrates the process. Specifically, we experiment with model merging approaches that inherently manage conflicts and mitigate modality dominance or instability when integrating modality-specific information using off-the-shelf LLMs, ensuring that neither modality overshadows the other.

\paragraph{Weight-based model merging:} We experiment with SLERP~\cite{goddard2024arcee}, which applies spherical interpolation to fuse model parameters in a more nuanced approach, blending models in a way that preserves unique characteristics. The goal is to find optimal coefficients $\lambda_1$ and $\lambda_2$ so that the merged model $\Theta^{(merge)} = \lambda_1 \tau_1 + \lambda_2 \tau_2$ retains the capabilities of the independent models. The coefficients are given by $ \frac{\sin((1-\lambda_1) \cdot \rho)}{\sin(\rho)} \quad \text{and} \quad \frac{\sin(\lambda_2 \cdot \rho)}{\sin(\rho)}$, respectively, where $\rho = \arccos \left( \frac{\tau_1 \cdot \tau_2}{|\tau_1| \cdot |\tau_2|} \right)$ is the angle between the task vectors, and $\lambda$ is the merging coefficient.

\paragraph{Subspace-based model merging:} We utilise TIES~\cite{yadav2023resolving} to prune the task vectors $\tau_1$ and $\tau_2$, retaining the top 20\% parameters, resulting in refined vectors $\hat{\tau_1}$ and $\hat{\tau_2}$ (see Fig.~\ref{fig:merging} (B)). We then fuse the vectors via Task Arithmetic~\cite{ilharco2022editing} to obtain the merged model as $\Theta^{(merge)} = \frac{1}{2} \sum_{i=1}^2 \hat{\tau_i}$. During the merging process, conflicts arising from differing signs in the parameters $p$ are resolved by aligning the pruned vectors as follows:

\begin{equation}
\text{Align}(\hat{\tau_1^p}, \hat{\tau_2^p}) = 
\begin{cases} 
\hat{\tau_1^p} & \text{if} \ |\hat{\tau_1^p}| > |\hat{\tau_2^p}| \\
\hat{\tau_2^p} & \text{if} \ |\hat{\tau_2^p}| \geq |\hat{\tau_1^p}|
\end{cases}
\end{equation}

\subsubsection{Crossmodal Alignment Fine-tuning}
\label{sec:alignment}
Let $\pi_{ref}$ be the reference policy (i.e., the universal model from model merging), $\pi_{\theta}$ the policy model being trained, parameterised by $\theta$, and $\mathcal{D}=\{x^{(i)},y_w^{(i)},y_l^{(i)}\}$ the offline preference data. Our goal is to learn effective crossmodals for the MoCG task with minimal training via alignment fine-tuning. We experiment with different optimizations that differ substantially in how they learn a reward signal, as overviewed in Table~\ref{tbl:alignment-methods}.
\begin{itemize}[noitemsep,topsep=0pt,parsep=0pt,partopsep=0pt,leftmargin=*] 
\item \textbf{SFT} minimises the difference between generated output \( z \) and target \( y_w \) by optimising model \( \pi_{\theta} \) through negative log-likelihood (Eq.~\ref{eq:sft}). 
\item \textbf{DPO}~\cite{rafailov2024direct} enhances cross-modal translations using an offline preference dataset \( \mathcal{D} \). It aligns model \( \pi_{\theta} \) by maximising the likelihood of preference data, with reference model \( \pi_{\text{ref}} \), Sigmoid function \( \sigma \), and hyperparameter \( \beta \) (Eq.~\ref{eq:dpo}).
\item  \textbf{CPO}~\cite{xu2024contrastive} reduces reliance on high-quality data by avoiding suboptimal translations, but not perfect translations in ML tasks.  It modifies Eq.~\ref{eq:dpo} using a uniform reference model, ensuring equal likelihood for all outputs. A behaviour cloning (BC) regulariser is injected to reflect uniform output matching, with an additional SFT term in the final loss (Eq.~\ref{eq:cpo}).

\begin{table}[h!]
\centering 
  \resizebox{\columnwidth}{!}{%
    \begin{tabular}{p{0.15\columnwidth}|p{0.85\columnwidth}}
        \toprule
        \textbf{Method} & \multicolumn{1}{c}{\textbf{Optimisation Objective}} \\ \hline
        SFT & 
        \vspace*{-\abovedisplayskip}
        \begin{align}
         \min_{\theta} -\log \pi_{\theta}(y_w|x) \label{eq:sft} 
        \end{align} \\ \hline

        DPO & 
        \vspace*{-\abovedisplayskip}
        \begin{align}
         \log \sigma \Bigl( \beta \log \frac{\pi_{\theta}(y_w|x)}{\pi_{\text{ref}}(y_w|x)} - \beta \log \frac{\pi_{\theta}(y_l|x)}{\pi_{\text{ref}}(y_l|x)} \Bigl) \label{eq:dpo} 
        \end{align} \\ \hline

        CPO & 
        \vspace*{-\abovedisplayskip}
        \begin{align}
         \min_{\theta} \log \sigma \Bigl( \beta \log \pi_{\theta}(y_w|x) - \beta \log \pi_{\theta}(y_l|x)\Bigl) - \log \pi_{\theta}(y_w|x) \quad \nonumber \\
        \text{s.t.} \quad \mathbb{E}_{(x,y_w)\sim D} \Bigl[ \mathbb{KL}\bigl(\pi_w(y_w|x) || \pi_{\theta}(y_w|x)\bigl) \Bigr] < \epsilon \label{eq:cpo}
        \end{align} \\ \hline

        KTO & 
        \vspace*{-\abovedisplayskip}
        \begin{align}
         -\lambda_{w} \sigma  \left( \beta \log \frac{\pi_{\theta}(y_w|x)}{\pi_{\text{ref}}(y_w|x)} - z_{ref}  \right) + \lambda_{l} \sigma \left( z_{ref} - \beta \log \frac{\pi_{\theta}(y_l|x)}{\pi_{\text{ref}}(y_l|x)}  \right) \nonumber \\
        \text{where } z_{\text{ref}} = \mathbb{E}_{(x,y) \sim \mathcal{D}} \left[ \beta \mathbb{KL}\bigl(\pi_{\theta}(y|x) \| \pi_{\text{ref}}(y|x)\bigr) \right] 
        \label{eq:kto}
        \end{align} \\ 
        \bottomrule
    \end{tabular}
    }
    \caption{Alignment fine-tuning algorithms for the MoCG task given preference data $\mathcal{D}=\{x,y_w,y_l\}$.}
    \label{tbl:alignment-methods}
\end{table}

\item \textbf{KTO}~\cite{ethayarajh2024kto} utilises non-paired preference data \( \mathcal{D} = \{x^{(i)}, y^{(i)}, \lambda^{(i)}\} \) where $\lambda$ denotes the desirability of \( y \). It directly maximizes the utility of generations instead of maximizing the log-likelihood of preferences. The loss is computed from the generated output \( z \) in relation to a reference \( z_{\text{ref}} \) and \( \lambda \) (Eq.~\ref{eq:kto}).
\end{itemize}

\subsection{Atomic-level Cross-NLI Evaluation}
\label{sec:atomic_nli}

\begin{figure*}[h!] 
  \centering
  \includegraphics[width=\textwidth]{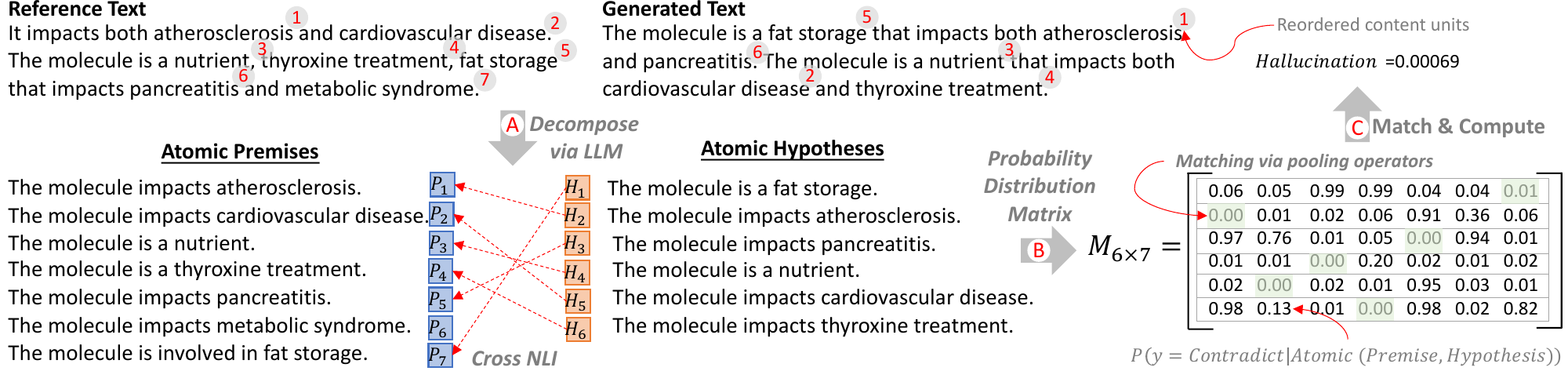} 
  \caption{The process of atomic-level cross-NLI evaluation when measuring the level of hallucination. }
  \label{fig:evaluation}
\end{figure*}

Our aim is to develop a method that operates at the right level of granularity, precisely capturing small distinctions and subtle nuances in captions, ensuring reliable evaluation. Atomic-level cross-NLI evaluation uses an LLM and an NLI model to assess relationships between generated and reference captions. The process begins with an LLM~\cite{touvron2023llama} decomposing a (reference, generated) pair into atomic premises $\{P_{i}\}_{i=1}^N$ and hypotheses $\{H_{j}\}_{j=1}^L$, where each atomic unit conveys a single piece of information (see Appx.~\ref{appx:instructions_eval}). An NLI model~\cite{he2020deberta} then constructs probabilistic distributions of entailment and contradiction by considering all possible combinations of premises and hypotheses. Finally, pooling operators match atomic hypotheses and premises in terms of both factual correctness, i.e.,  \textit{hallucination}, and completeness, i.e., \textit{coverage}. Fig.~\ref{fig:evaluation} illustrates this process.

\noindent \textbf{Hallucination} we define here as the introduction of information not present in the reference text. Given $\{(P_i, H_j)\}$, the NLI model constructs a contradiction probability distribution for each atomic hypothesis against all premises, such as $p_{j,i}=(C_{j,i} | P_i, H_j)$. This results in an $M_{L \times N}$ matrix of contradiction probabilities $C_{j,i}$ (see Fig.~\ref{fig:evaluation}). To measure hallucination, we apply min row-wise pooling and average the matching probabilities to compute the score by the formula:

\begin{equation}
    Hallucination = \frac{1}{L} \sum_{j=1}^{L} \min_{i} C_{j,i}
\end{equation}

\paragraph{Coverage} we define as atomic unit recall, representing how much reference information is present in the generated text. Unlike hallucination, here generated text forms the atomic premises ($P_j$) and the reference text the hypotheses ($H_i$). The NLI model constructs an entailment probability distribution for each $H_i$ against all $P_j$, such that $p_{i,j}=(E_{i,j} | P_j, H_i)$, resulting in an $M_{N \times L}$ matrix of entailment probabilities $E_{i,j}$. To measure coverage, we apply max row-wise pooling and average the matching probabilities to compute the score given by the formula:

\begin{equation}
    Coverage = \frac{1}{N} \sum_{i=1}^{N} \max_{j} E_{i,j}
\end{equation}
\section{Experiments}

\subsection{Experimental Setup}
\label{sec:setup}
\paragraph{Data:}We conduct experiments training Meditron~\cite{chen2023meditron} on the benchmark  L+M-24~\cite{edwards2024l+} dataset, using only 10\% of the data for training, and evaluate on out-of-distribution data (see Appx.~\ref{appx:data} for details). For alignment fine-tuning, we create synthetic dispreferred outputs generated by MolT5~\cite{edwards2022translation}. In practice, this involves feeding MolT5 with inputs from the 10\% subset of L+M-24 used in our experiments, generating outputs, and then using these outputs as dispreferred samples (see Fig.~\ref{fig:framework} ). Our training, validation, and test sets contain approximately 12.7k, 3.4k, and 3k samples. 

\paragraph{Baselines:} We selected established baselines based on their relevance to our hypotheses, enabling comparison with models trained on fully (i.e., Chem-LLM~\cite{zhang2024chemllm}) and partially (i.e., TxtChem-T5~\cite{christofidellis2023unifying}) out-of-distribution data, as well as in-distribution data (Meditron~\cite{chen2023meditron}). In this context, TxtChem-T5 and Chem-LLM are evaluated in a zero-shot setting. For more details about the baselines, please refer to Appx.~\ref{appx:baselines}.
Lastly, we fine-tune Meditron with \textit{SFT}  using only 10\% of the training data. We leave all the implementation details in Appx.~\ref{appx:implementation}.

\paragraph{Evaluation:} When evaluating the performance of both baselines and our models, we employ established statistical metrics (see Appendix~\ref{appx:metrics}), in addition to our atomic-level cross-NLI evaluation method (\S~\ref{sec:atomic_nli}). For our proposed evaluation, we assess the robustness of different NLI methods by measuring the relative entropy of textual entailment between generated outputs from high and low performance models in association with linguistic ones derived by bioinformatic databases curated by humans. Specifically, we compare our atomic-level NLI approach with leading ones, including \textit{full NLI}, which treats entire premises and hypotheses as single units, and \textit{sentence-level NLI}~\cite{laban2022summac}, i.e., which evaluates chunks in text.

\subsection{Experimental Results}

\subsubsection{Aligning Molecule-Language Modals with Minimal Training}
\label{sec:alignemt_results}
We first present results for molecule language models with minimal alignment fine-tuning, initialising pretrained weights from molecule generation rather than deploying model merging (see Appx.~\ref{appx:implementation} for details). Tables~\ref{tbl:mol2cap} and \ref{tbl:cap2mol} summarise experimental results. Generally, benchmarking models trained on extensive data with SFT exhibit memorisation effects, with performance dropping by 50\% to 100\% compared to reported results, when evaluated on out-of-distribution data.

Our experiments show that not all alignment optimisations are effective in the minimal training setting. Both DPO and KTO show zero performance in caption generation when models are initialised with crossmodal weights unrelated to the task (see Table~\ref{tbl:mol2cap}). However, performance improves significantly when the crossmodals are known (see Table~\ref{tbl:cap2mol}). In molecule generation, DPO achieves up to 42\% better performance than Meditron, trained on the full dataset, while KTO still performs poorly, likely due to overfitting (see Appx.~\ref{appx:efficiency}).

By contrast, CPO effectively handles both the crossmodal agnostic and minimal training settings, outperforming Meditron by up to 20\% in caption generation and 42\% in molecule generation. This is likely due to its inherent ability to balance structured learning and generalisation. It aligns with preferred data through behaviour cloning and SFT, which encourage the model to mimic expert behaviour while reducing bias and suboptimal outcomes via a uniform reference model that assigns equal likelihood to all possible outputs.

\begin{table*}[ht]
  \centering 
  \resizebox{\textwidth}{!}{%
  \begin{tabular}{ccccccc }
    \toprule

    \textbf{Method} & $\textbf{Blue-2} \uparrow$ & $\textbf{Blue-4} \uparrow$ & $\textbf{Rouge-1} \uparrow$  &   $\textbf{Rouge-2} \uparrow $ &  $\textbf{Rouge-L} \uparrow$  & $\textbf{METEOR} \uparrow$     \\ \hline

    TxtChem-T5~\cite{christofidellis2023unifying} & 0.08  & 0.09  &  0.19  & 0.06 &  0.17 & 0.16\\ \hline
    Chem-LLM~\cite{zhang2024chemllm} & 0.03 & 0.00  &  0.11  & 0.02 & 0.09 & 0.14 \\ \hline
    Meditron~\cite{chen2023meditron} &  0.42 &  0.30 &  0.63  & 0.47 & \textbf{0.49} & 0.54\\  \hline \hline
    SFT~$\S\ref{sec:setup}$ &  0.37 & 0.26 & 0.55  & 0.40& 0.39 & 0.61\\  \hline 
    DPO~\cite{rafailov2024direct} & 0.00  & 0.00 & 0.00  & 0.00& 0.00 & 0.00\\  \hline 
    \textbf{CPO}~\cite{xu2024contrastive} & \textbf{0.62}  & \textbf{0.45} & \textbf{0.68}  &\textbf{0.50}& 0.48 & \textbf{0.62}\\  \hline 
    KTO~\cite{ethayarajh2024kto} & 0.00  & 0.00 & 0.00  & 0.00 & 0.00 & 0.00\\  \hline  \hline
    $\Delta_{CPO \text{vs} MED }$ & \textcolor{Green}{+20\%}  & \textcolor{Green}{+19\%}  & \textcolor{Green}{+5\%}  & \textcolor{Green}{+3\%}& \textcolor{Red}{-1\%} & \textcolor{Green}{+8\%} \\  
    
  \bottomrule
  \end{tabular}
 }
  \caption{Alignment fine-tuning results for caption generation on  3k unseen pairs. Arrows next to metrics denote value increase with performance gains. Best results are in bold. $\Delta_{CPO \text{vs} MED }$ is the performance gain of our best model, trained on 10\% of the data, compared to Meditron trained on the entire dataset.}
  \label{tbl:mol2cap}
\end{table*}

\begin{table*}[ht]
  \centering
 
  \resizebox{\textwidth}{!}{%
  \begin{tabular}{cccccccc }
    \toprule
    \textbf{Method} & $\textbf{BLEU} \uparrow$  & $\textbf{Levenshtein} \downarrow$  &   $\textbf{MACCS FTS} \uparrow $ &  $\textbf{RDK FTS} \uparrow$  & $\textbf{Morgan FTS} \uparrow$ & $\textbf{FCD} \downarrow$ &    $\textbf{Validity} \uparrow$  \\ \hline

    TxtChem-T5 & 0.18  &  133.29  & 0.21 & 0.10 & 0.03 &  37.67 & 0.58\\ \hline
    Chem-LLM  &  0.04   &  732.74 &  0.00 &  0.00  & 0.00   & 59.44  & 0.19 \\ \hline
    Meditron & 0.43   &  66.16  & 0.35 & 0.29  & 0.19 & 13.64 & 0.57\\ \hline \hline
    SFT & 0.30   & 186.99   & 0.70  & 0.62  & 0.41 & 11.14 & 0.98 \\ \hline
    \textbf{DPO} & \textbf{0.72}   &  \textbf{42.40}  & \textbf{0.77}  & 0.69  & \textbf{0.49} & 10.47 &  0.99 \\ \hline
    \textbf{CPO} & 0.71   & 42.65   &  \textbf{0.77}  & \textbf{0.70}  & 0.48 & \textbf{4.19} &  \textbf{1.00}\\ \hline
    KTO & 0.23    & 294.63   & 0.03  & 0.03  & 0.02 & 32.64 & 0.06  \\ \hline \hline
    $\Delta_{CPO \text{vs} MED }$  &  \textcolor{Green}{+29\%}     &  \textcolor{Green}{-23.76\%}    & \textcolor{Green}{+42\%}  & \textcolor{Green}{+41\%}   & \textcolor{Green}{+30\%}    &\textcolor{Green}{-9.45\%}    & \textcolor{Green}{+41\%}  \\ 
  \bottomrule
  \end{tabular}
 }
  \caption{Alignment fine-tuning results for molecule generation on 3k unseen pairs. Arrows next to metrics indicate whether higher or lower values denote better performance. Best results are highlighted in bold. $\Delta_{CPO \text{vs} MED }$ represents the performance gain of our best model compared to Meditron trained on the entire dataset.}
  \label{tbl:cap2mol}
\end{table*}

\subsubsection{Alignment with Model Merging}
\label{sec:merging}
Tables~\ref{tbl:mol2cap_fusion} and \ref{tbl:cap2mol_fusion} summarise the experimental results when we incorporate model merging in alignment fine-tuning while keeping the training data the same. 
Combining DPO with molecule and caption crossmodals via TIES improves caption generation (see $\Delta_{DPO \text{vs} TIES+DPO}$ in Table~\ref{tbl:mol2cap_fusion}) but leads to significant performance loss in molecule generation (see $\Delta_{DPO \text{vs} TIES+DPO}$ in Table~\ref{tbl:cap2mol_fusion}). Conversely, fusing CPO with crossmodals via SLERP significantly boosts performance in caption generation (see $\Delta_{CPO \text{vs} SLERP+CPO}$ in Table~\ref{tbl:mol2cap_fusion}) while having minimal impact on molecule generation (see $\Delta_{CPO \text{vs} SLERP+CPO}$ in Table~\ref{tbl:cap2mol_fusion}), demonstrating overall gains compared to Meditron trained on the full dataset.

\begin{figure}[h!]
\centering
\includegraphics[width=1\columnwidth]{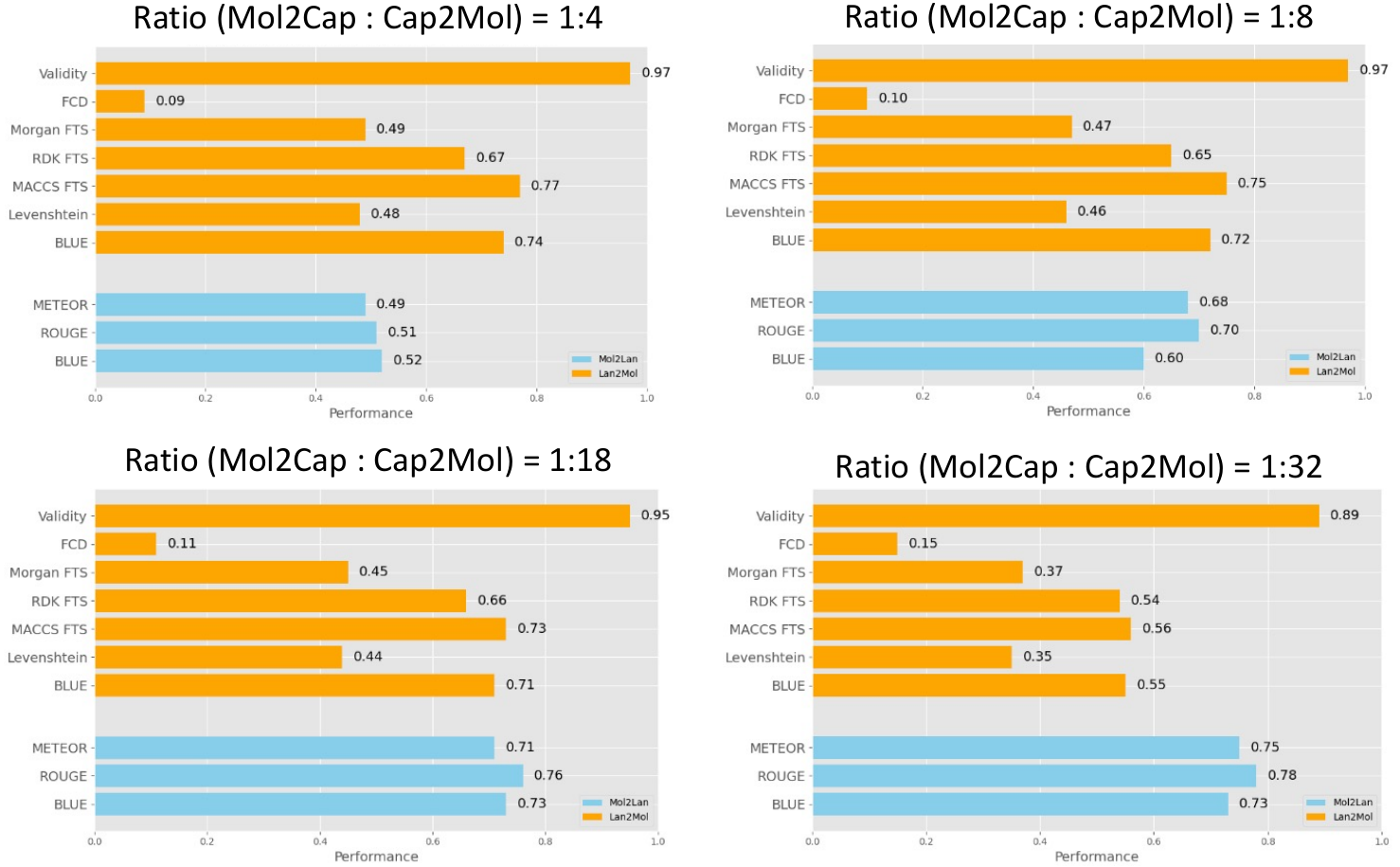}
\caption{Ablation of best performance model, CPO+SLERP, for Mol2Cap and Cap2Mol tasks, evaluating the effect of per-task model weight mixing ratios.}
\label{fig:ablation_merging}
\end{figure}

For our best-performing model, CPO+SLERP, we conducted ablation studies to assess the impact of weight interpolation coefficients when merging pretrained models on MoCG tasks. Specifically, we explored blending weights across all layers (0–32) to preserve Mol2Cap performance while improving Cap2Mol performance (see Appx.~\ref{appx:mergin_abblations} for details), aiming to create a universal model with enhanced overall capability. Fig.~\ref{fig:ablation_merging} shows the performance trends across different mixing ratios of per-task model weights. Empirically, we found that a 1:18 ratio (Mol2Cap:Cap2Mol) yields the best balance, favoring Mol2Cap performance at lower ratios and Cap2Mol performance at higher ones.  Further comparison with a baseline method, namely model soup~\cite{wortsman2022model}, is provided in Appx.~\ref{appx:mergin_abblations}.

Overall, our experiments show that model merging mitigates key limitations in alignment fine-tuning. By fusing pretrained models, it boosts performance with minimal training, reduces reliance on human-labeled data, lowers costs, minimizes bias, and improves generalization. Caption and molecule generation examples are in Appx.~\ref{appx:examples_gen}.

\begin{table*}[ht]
   \centering
   \resizebox{\textwidth}{!}{%
     \begin{tabular}{cccccccc}
       \toprule
       \textbf{Fusion} & \textbf{Method} & \textbf{Blue-2} $\uparrow$ & \textbf{Blue-4} $\uparrow$ & \textbf{Rouge-1} $\uparrow$ & \textbf{Rouge-2} $\uparrow$ & \textbf{Rouge-L} $\uparrow$ & \textbf{METEOR} $\uparrow$ \\ \midrule

       \multirow{2}{*}{TIES~\cite{yadav2023resolving} } & DPO &0.74 & 0.53& 0.74 & 0.54& 0.51 & 0.70 \\ \cline{2-8} 
                              & CPO &0.74 & 0.54 & 0.76 & 0.57& 0.53 & 0.72 \\ \bottomrule

      \multirow{2}{*}{SLERP~\cite{goddard2024arcee}} & DPO & 0.00 & 0.00& 0.02 &0.01 & 0.00 & 0.00 \\ \cline{2-8} 
                              & CPO & 0.73 & 0.53 & 0.76 & 0.56 & 0.53 & 0.71\\ \hline  \hline 
      \multicolumn{2}{c}{$\Delta_{DPO \text{vs} TIES+DPO }$}     & \textcolor{Green}{+74\%}  & \textcolor{Green}{+53\%}  & \textcolor{Green}{+74\%}   & \textcolor{Green}{+54\%} & \textcolor{Green}{+51\%}  & \textcolor{Green}{+70\%}  \\   \hline
       \multicolumn{2}{c}{$\Delta_{CPO \text{vs} SLERP+CPO }$} &  \textcolor{Green}{+11\%}  &  \textcolor{Green}{+8\%}   &\textcolor{Green}{+8\%}  & \textcolor{Green}{+6\%}& \textcolor{Green}{+5\%} & \textcolor{Green}{+9\%}  \\  \hline
       \multicolumn{2}{c}{$\Delta_{MED \text{vs} SLERP+CPO }$}   & \textcolor{Green}{+31\%}  & \textcolor{Green}{+28\%}  & \textcolor{Green}{+13\%} & \textcolor{Green}{+9\%} & \textcolor{Green}{+4\%} & \textcolor{Green}{+17\%} \\   
       
                              \bottomrule                        
     \end{tabular}
   }
   \caption{Model merging and alignment fine-tuning results for caption generation. $\Delta_{DPO \text{vs} TIES+DPO }$, $\Delta_{CPO \text{vs} SLERP+CPO }$, and $\Delta_{MED \text{vs} SLERP+CPO }$ measure performance gains of the best-combined approaches compared to the vanilla crossmodal setting of \textit{DPO}, \textit{CPO}, and the benchmark \textit{Meditron}, as reported in Table~\ref{tbl:mol2cap}. }
   \label{tbl:mol2cap_fusion}
 \end{table*}

  \begin{table*}[ht]
   \centering
   \resizebox{\textwidth}{!}{%
     \begin{tabular}{ccccccccc}
       \toprule
       \textbf{Fusion} & \textbf{Method} & $\textbf{BLEU} \uparrow$  & $\textbf{Levenshtein} \downarrow$  &   $\textbf{MACCS FTS} \uparrow $ &  $\textbf{RDK FTS} \uparrow$  & $\textbf{Morgan FTS} \uparrow$ & $\textbf{FCD} \downarrow$ &    $\textbf{Validity} \uparrow$  \\ \midrule
       \multirow{2}{*}{TIES} & DPO  &0.32 & 93.18 & 0.31 & 0.22& 0.19&  19.80 & 0.42  \\ \cline{2-9} 
                              & CPO & 0.68& 46.91 & 0.72& 0.65 & 0.45 & 24.50 &0.94\\ \bottomrule

      \multirow{2}{*}{SLERP} & DPO & 0.72& 43.85 & 0.77& 0.70&  0.51 & 10.35 & 0.98\\ \cline{2-9} 
                              & CPO & 0.71 & 44.01 & 0.73 & 0.66& 0.45 & 11.22 &  0.95\\ \hline  \hline 
          \multicolumn{2}{c}{$\Delta_{DPO \text{vs} TIES+DPO }$}   & \textcolor{Red}{-40\%}  & \textcolor{Red}{+51\%} & \textcolor{Red}{-46\%}& \textcolor{Red}{-47\%}  &\textcolor{Red}{-30\%}  & \textcolor{Red}{+7.33\%} & \textcolor{Red}{+58\%} \\   \hline
      \multicolumn{2}{c}{$\Delta_{CPO \text{vs} SLERP+CPO }$}  &  \textcolor{Green}{0\%} &\textcolor{Red}{+1.36\%}  & \textcolor{Red}{-4\%} & \textcolor{Red}{-4\%} & \textcolor{Red}{-3\%} & \textcolor{Red}{+5\%} & \textcolor{Red}{-4\%} \\  \hline
       \multicolumn{2}{c}{$\Delta_{MED \text{vs} SLERP+CPO }$}    & \textcolor{Green}{+29\%}  & \textcolor{Green}{-22.40\%} &  \textcolor{Green}{+38\%} &  \textcolor{Green}{+37\%}  &  \textcolor{Green}{+27\%} & \textcolor{Green}{-4.45\%}  & \textcolor{Green}{+37\%}  \\                           
                              \bottomrule                        
     \end{tabular}
   }
   \caption{Model merging and alignment fine-tuning results for molecule generation. $\Delta_{DPO \text{vs} TIES+DPO }$, $\Delta_{CPO \text{vs} SLERP+CPO }$, and $\Delta_{MED \text{vs} SLERP+CPO }$ measure performance gains of the best-combined approaches from the vanilla crossmodal setting of \textit{DPO}, \textit{CPO}, and the benchmark \textit{Meditron}, as reported in Table~\ref{tbl:mol2cap}.}
   \label{tbl:cap2mol_fusion}
 \end{table*}

\subsubsection{Atomic-level Cross-NLI Evaluation}
\label{sec:atomic_results}
\begin{figure}[h!]
\centering
\includegraphics[width=1\columnwidth]{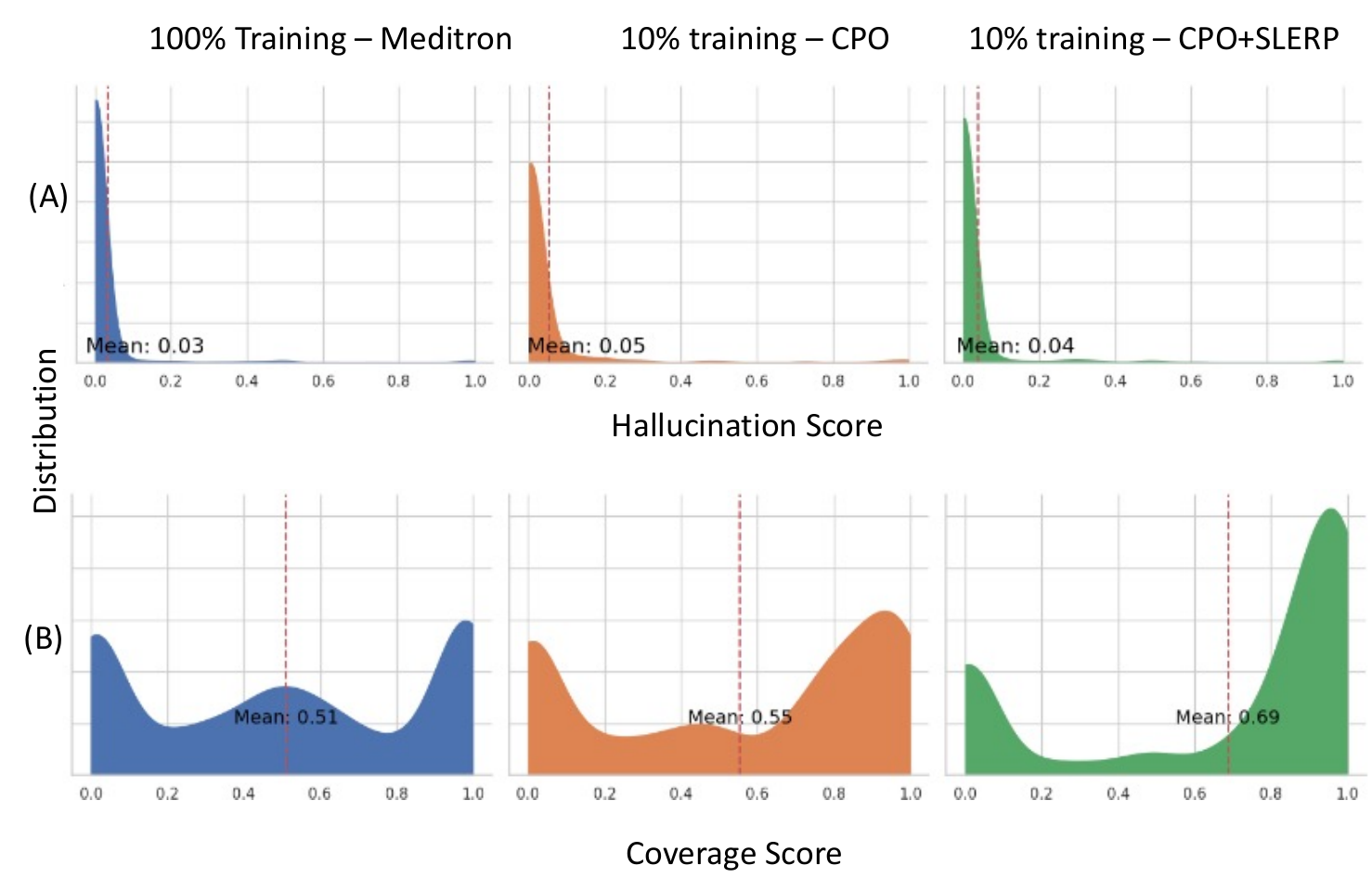}
\caption{Score distributions from our atomic-level cross-NLI evaluation comparing (A) hallucination and (B) coverage between our top models and Meditron.}
\label{fig:atomic_evaluation}
\end{figure}

Atomic-level NLI revealed intriguing insights regarding performance interpretation. Fig.~\ref{fig:atomic_evaluation} shows assessment score distributions from our proposed evaluation method, comparing our top models against Meditron trained on the entire dataset. All models exhibit low hallucination, likely due to the narrow, well-defined topics that enable factually correct captions without unrelated information. However, our models excel in coverage, generating more comprehensive captions, with performance increasing to 69\% compared to Meditron's 51\% (Fig.~\ref{fig:atomic_evaluation} (B)). Examples of insights captured by our proposed evaluation are in  Appx.~\ref{appx:nli_cases}.

\begin{figure}[h!]
\centering
\includegraphics[width=1\columnwidth]{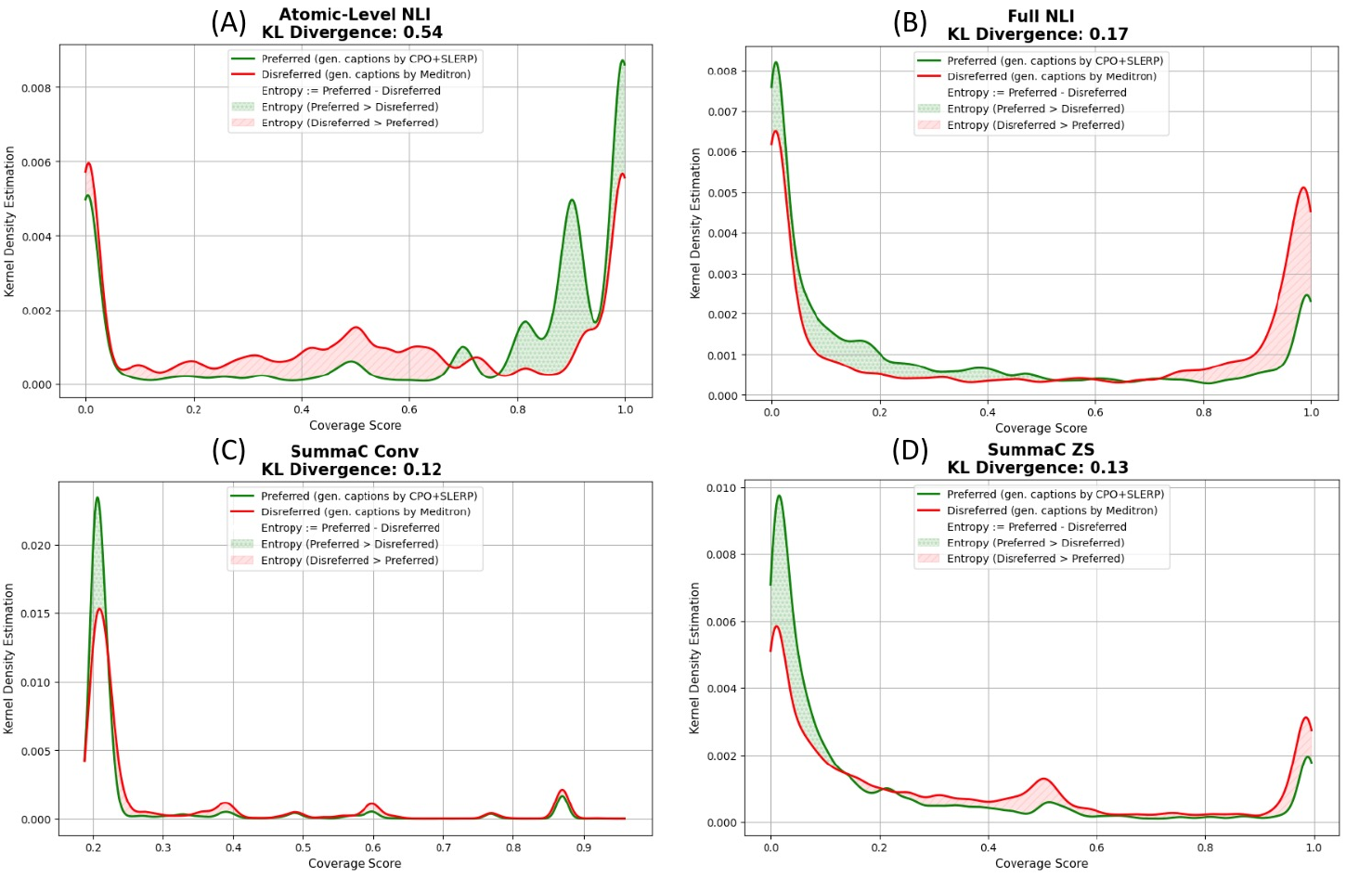}
\caption{Relative entropy in coverage scores for preferred vs. dispreferred generated captions across atomic-level (A), full (B), and sentence-level (C \& D) NLI approaches.}
\label{fig:entropy}
\end{figure}

We also evaluated the robustness of our proposed NLI evaluation method against leading approaches by measuring the relative entropy of textual entailment between human-curated texts (i.e., gold labels) and outputs generated by our top-performing model, CPO+SLERP (preferred), versus those from a low-performing model, Meditron (dispreferred). Ideally, all NLI methods should favour preferred outputs over dispreferred ones. However, we observed that both the full and sentence-level NLI methods misclassify preferred captions as non-entailment and dispreferred captions as entailment (see Fig.~\ref{fig:entropy} (B)-(D)). By contrast, atomic-level cross-NLI accurately favours preferred captions, assigning higher scores to certain cases (Fig.~\ref{fig:entropy} (A)). Additionally, Kullback–Leibler divergence shows that atomic-level NLI offers better discrimination, achieving a divergence score of 0.54 compared to 0.12–0.17 for other methods, demonstrating its effectiveness in distinguishing the quality of generated captions. We leave further ablation analysis in Appx.~\ref{appx:nli_abblations}.

\begin{figure}[h!]
\centering
\includegraphics[width=1\columnwidth]{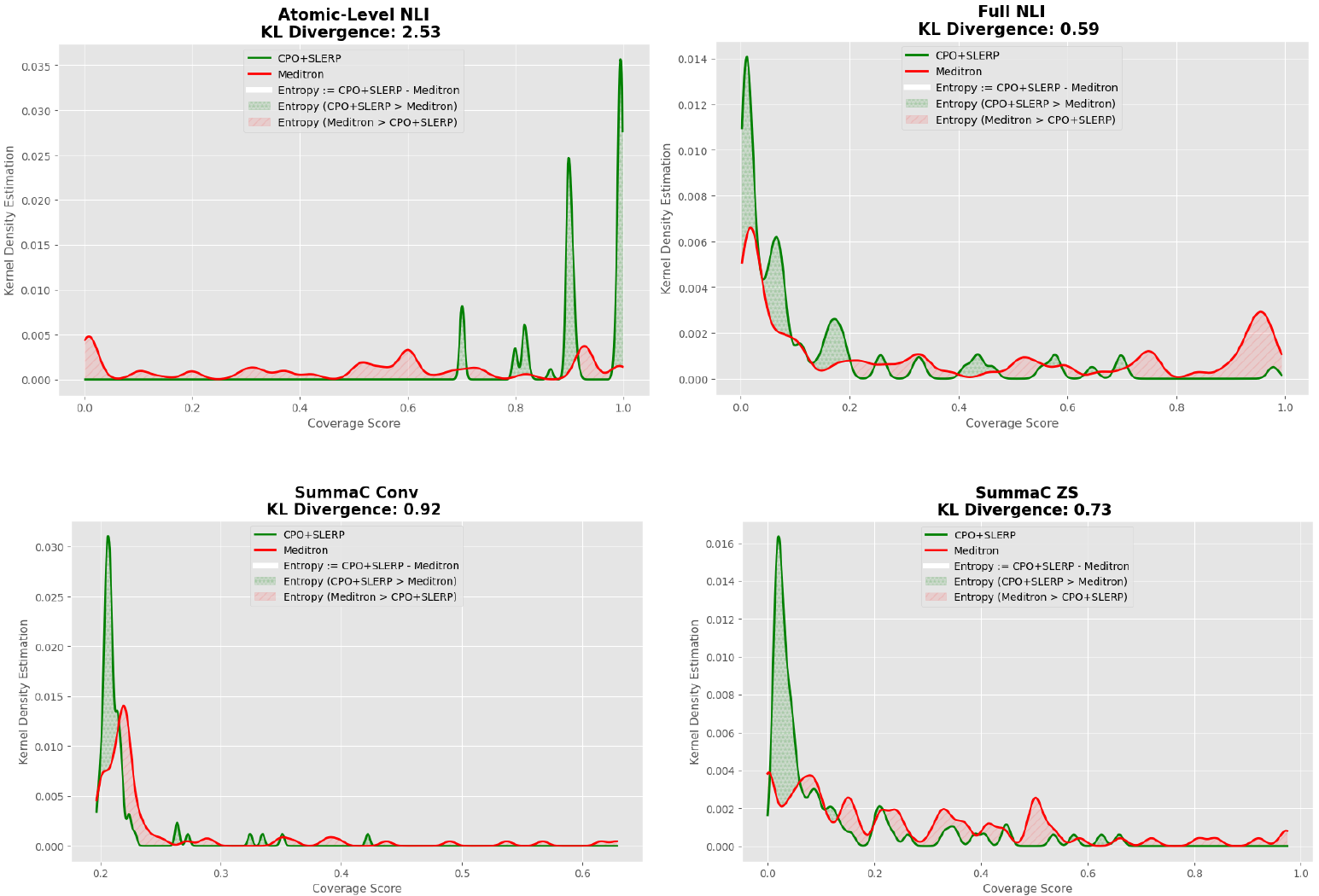}
\caption{Relative entropy in coverage scores for preferred vs. dispreferred generated captions across atomic-level and leading NLI approaches in long captions.}
\label{fig:nl1}
\end{figure}

We conducted ablation studies on our atomic-level NLI evaluation method to assess its semantic robustness, particularly in handling complex, lengthy captions that may lose cohesiveness due to excessive decomposition into atomic units. To evaluate this, we analyzed the word count distribution (see Appx.~\ref{appx:nli_abblations}), filtered captions with at least 50 words, and recalculated relative entropy against standard NLI methods. Fig.~\ref{fig:nl1} demonstrates the superior performance of our method on longer cases compared to leading alternatives. Our NLI method demonstrated a significant improvement in its ability to differentiate preferred outputs from dispreferred ones accurately, achieving a KL divergence of 2.53 (see Fig. \ref{fig:nl1}), as opposed to a KL divergence of 0.54 across all cases in the test subset (see Fig. \ref{fig:entropy}). In contrast, other leading NLI methods experienced a marked increase in KL divergence, favouring dispreferred outputs, which misaligned with the entailment aspect. 

\section{Conclusion}
In this work, we address limitations of scientific language models that rely on extensive training. Focusing on molecule caption generation, we propose synergies between model merging and alignment fine-tuning with minimal post-training to enhance chemical language models. Our experiments show that while alignment fine-tuning performs poorly, incorporating model merging significantly outperforms extensively trained models on out-of-distribution data, offering a cost-effective approach that relies less on human-labelled data. Furthermore, we propose an atomic-level cross-NLI evaluation to overcome limitations of widely used NLI evaluation methods, which lack appropriate granularity. Our method provides valuable insight into performance interpretability and effectively handles multiple content units, where existing NLI methods consistently misalign with assessment criteria.

\section*{Limitations}
In this work, we employ weight-based and subspace-based merging methods to create universal models for the MoCG task, facilitating alignment fine-tuning in a training setting with minimal data. However, both are static merging methods. This means the merged model remains the same for all samples or tasks. Given that there are differences between input samples/tasks, the models' ability may vary when processing different samples/tasks. In the future, we aim to investigate dynamically merging models (or subsets of layers) based on the samples/tasks during the inference phase~\cite{kang2024self,yang2024model}.

We also propose an atomic-level NLI evaluation method that successfully handles multiple content units, offering valuable insights into performance interpretability for caption generation, where widely adopted NLI methods consistently misalign with assessment criteria. However, decomposing text into atomic units can be challenging for other tasks involving complex or lengthy text. While this method captures nuanced content, there is a risk of over-fragmentation, which may lead to a loss of context or coherence in evaluation. Additionally, the effectiveness of this approach relies heavily on the LLM for decomposition and the NLI model for entailment and contradiction assessment. The evaluation could yield inaccurate or biased results if either model struggles with domain-specific content (e.g., highly technical language). Furthermore, if generated texts introduce valid but creative or non-standard content, this approach may penalise them by classifying such deviations as contradictions or hallucinations, even when they provide accurate information. Future work will need to address these limitations across various domains.

Finally, the proposed methods in this work are tailored specifically for the chemical domain, focusing on tasks like molecule caption generation. While these techniques—such as model merging and alignment fine-tuning—show promising results within this context, their ability to generalise to other domains or scientific fields is uncertain. Different domains may have distinct data structures, tasks, and requirements, which might not align well with the crossmodal setup used here. For instance, a method optimised for chemical language and molecular structures may not work as effectively in domains like physics or biology, where the types of entities and relationships differ significantly.  This potential lack of generalisation highlights the need for future research to explore the applicability of the proposed approaches in diverse scientific domains beyond chemistry, aiming to adapt and validate the methods for varying data structures and task requirements.

\section*{Ethical Considerations}
The potential for generating misleading or incorrect information poses significant ethical considerations in this work, particularly given the scientific context in which the language models are applied. If the models produce inaccurate captions or misrepresent molecular characteristics, it could lead to erroneous conclusions in research and applications that rely on these outputs. This risk is particularly critical in fields like chemistry, where precise data interpretation is vital for safety, compliance, and advancing scientific knowledge. Furthermore, the reliance on automated evaluations may not adequately catch nuanced errors that human experts would recognise, potentially allowing flawed outputs to go unchecked. Therefore, ensuring that the models maintain a high standard of accuracy and reliability is essential to prevent the dissemination of misinformation, which could undermine trust in automated systems and hinder scientific progress. Addressing these ethical concerns requires implementing robust validation mechanisms and continuously involving domain experts in the evaluation process to ensure the integrity of the generated content.

\section*{Acknowledgements}
This work was supported by a UKRI/EPSRC Turing AI Fellowship to Maria Liakata (grant ref EP/V030302/1) and the Alan Turing Institute (grant ref EP/N510129/1). This work was supported by the Engineering and Physical Sciences Research Council [grant number EP/Y009800/1], through funding from Responsible Ai UK (KP0016) as a Keystone project lead by Maria Liakata.
\bibliography{custom}
\clearpage

\appendix
\section{Complementary Experiments in Model Merging}
\label{appx:mergin_abblations}
\begin{figure}[h!]
\centering
\includegraphics[width=1\columnwidth]{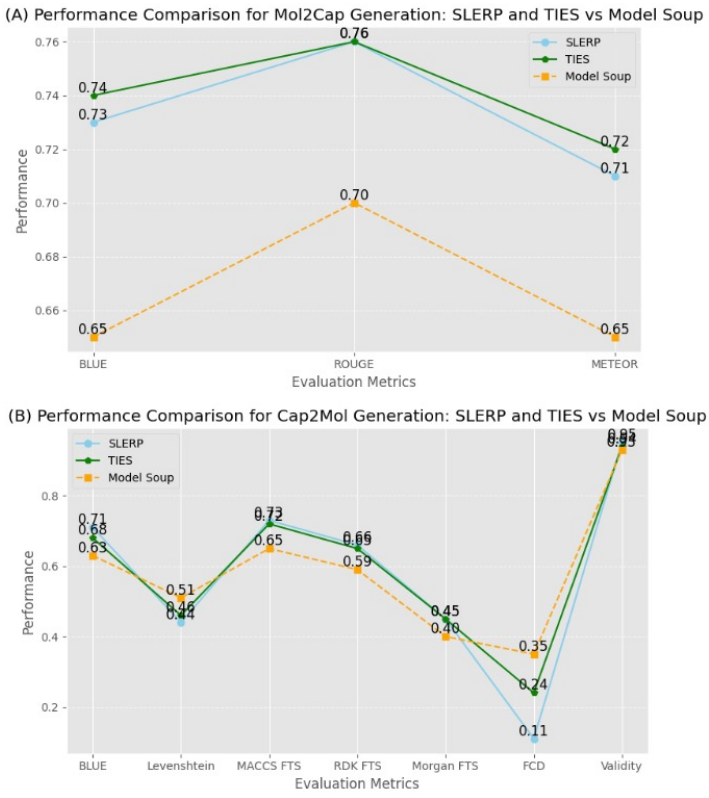}
\caption{Comparison of SLERP and TIES with Model Soup for (A) Mol2Cap and (B) Cap2Mol generation. }
\label{fig:model_soup}
\end{figure}

We compared SLERP and TIER model merging techniques against a weighted linear combination of parameters, referred to as model soup~\cite{wortsman2022model}, when applying CPO in the MoCG task. Our results indicated that model soup caused a significant drop in performance for both Mol2Cap and Cap2Mol tasks (see Fig.~\ref{fig:model_soup}). We hypothesise that this is because model soup assumes that performance improvement or preservation is linearly related to weight blending, which may not hold for complex models. This observation justifies our decision to explore task-specific arithmetic and geometric merging approaches, as they inherently manage conflicts and better preserve the strengths of each model in specialised tasks.

\section{Complementary Experiments in Our Atomic-Level NLI Evaluation Method}
\label{appx:nli_abblations}

\begin{figure}[h!]
\centering
\includegraphics[width=1\columnwidth]{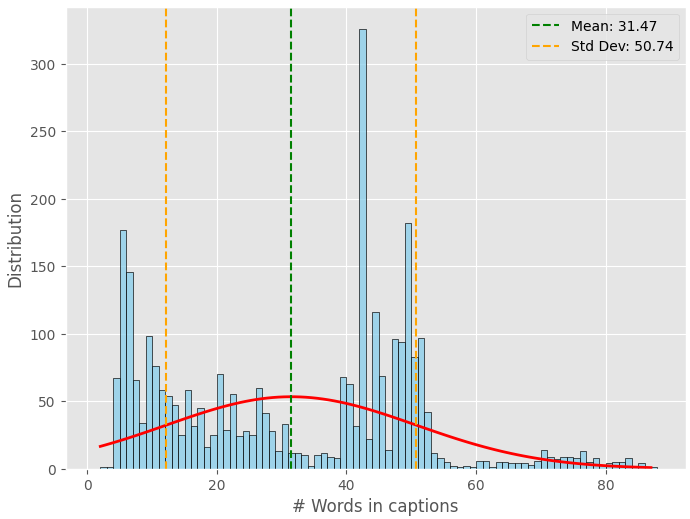}
\caption{Distribution of word counts in captions from the test subset.}
\label{fig:distributions}
\end{figure}

First, we analysed the distribution of word counts in captions from the test subset. We observed that the captions are typically short, with an average of 31 words (STD = 50) as shown in Fig.\ref{fig:distributions}. Additionally, the captions generally exhibit little dependency across sentences, as they consist of simple natural language describing chemical properties (for a more detailed view, see Table~\ref{tbl:nli_examples}).

\begin{figure}[h!]
\centering
\includegraphics[width=1\columnwidth]{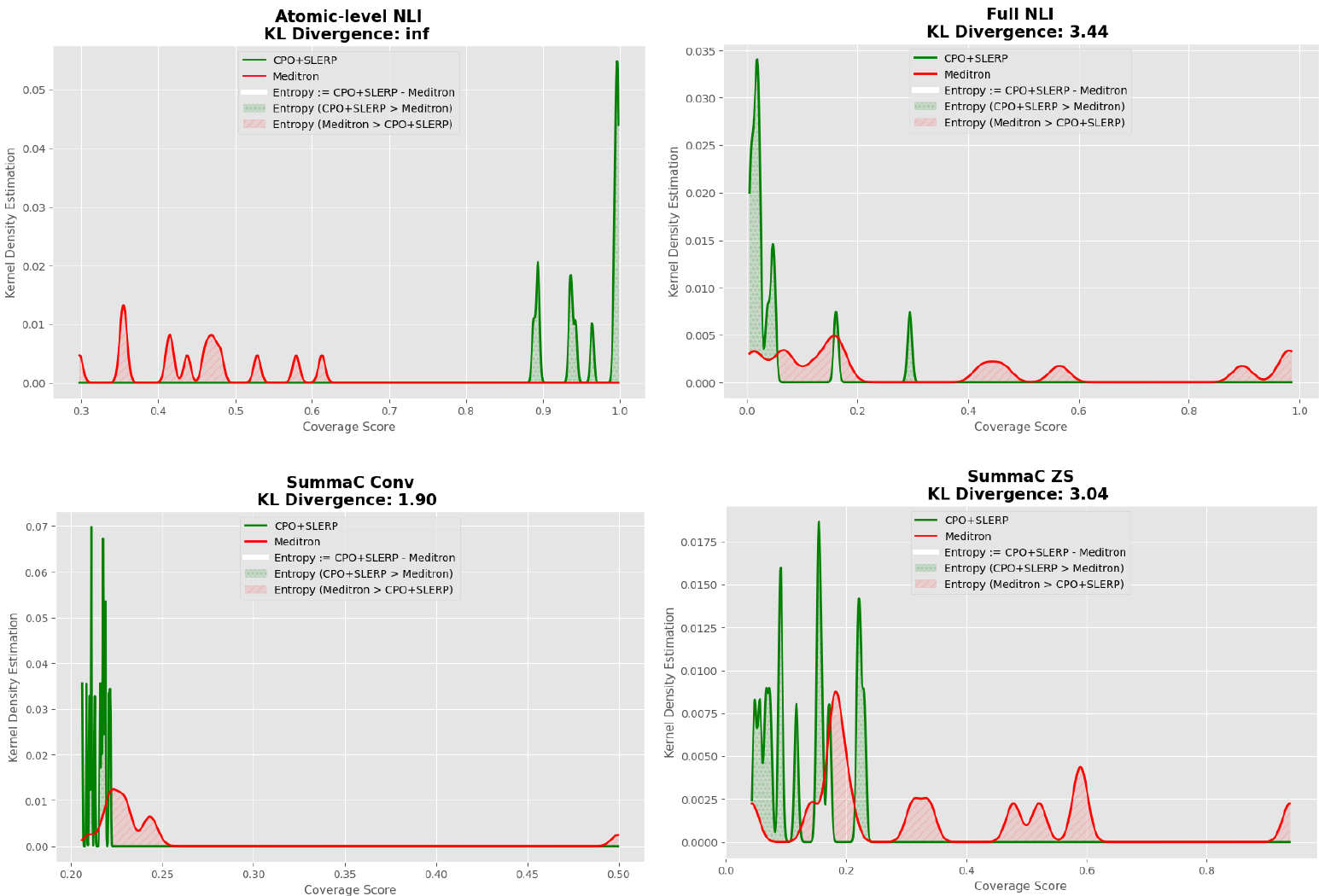}
\caption{Relative entropy in coverage scores for preferred vs. dispreferred generated captions across atomic-level and leading NLI approaches in extreme captions.}
\label{fig:nl2}
\end{figure}

Based on the word count distribution analysis, we filtered captions with at least 70 words and recalculated the relative entropy against standard NLI methods. As shown in Fig.~\ref{fig:nl2}, our method demonstrates superior performance on extremely long cases compared to leading alternatives. Notably, the performance trend is consistent with that observed for generally lengthy captions (at least 50 words, see $\S~\ref{sec:atomic_results}$).

\section{Foundations in Alignment with RLHF}
\label{appx:foundations_rlhf}

Feedback-aligned LLMs traditionally undergo fine-tuning with RLHF, where human preferences serve as a reward signal in optimisation~\cite{stiennon2020learning,ouyang2022training}. To train a LLM with RLHF, a reinforcement learning optimisation algorithm such as PPO~\cite{schulman2017proximal} is typically deployed on offline preference data, commonly involving three steps:

\begin{itemize}[noitemsep,topsep=0pt,parsep=0pt,partopsep=0pt,leftmargin=*]

\item \textbf{Model Training:} Typically, a model $\pi$ is trained for auto-regressive language generation on a large generic corpus. This training operates under the premise that the probability distribution of a sequence of words can be broken down into the product of conditional distributions for the next word~\cite{radford2019language}.

\item  \textbf {Reward Model Training:} A reference model  $\pi_{\text{ref}}$ is employed to optimise $\pi$ for a downstream task. Typically, the $\pi_{\text{ref}}$ model undergoes fine-tuning with an auto-regressive objective, using data pertinent to the downstream task. This often involves instruction tuning $\pi_{\text{ref}}$ to regulate the generated outputs.

\item  \textbf {Reinforcement Learning:}
The optimisation of $\pi$ with respect to $\pi_{\text{ref}}$ operates on a triple dataset $\mathcal{D}= \{x, y_w, y_l\}$, where $x$ represents the input, and $y_w$ and $y_l$ denote preferred and dis-preferred outputs, respectively, such that $y_w \succ y_l$ for $x$. In the Bradley–Terry model~\cite{bradley1952rank}, the probability of $y_w$ being preferred over $y_l$ in pairwise comparisons can be formulated as follows:
\begin{equation}
    p^* (y_w \succ y_l | x) = \sigma (r^*(x,y_w)-r^*(x,y_l))
\end{equation}
Here, $\sigma$ represents the logistic function, and $r^*$ denotes the ``true'' reward function that underlies the preferences. As obtaining the true reward directly from a human would be prohibitively expensive, a reward model $r_\phi$ is trained to act as a surrogate. This is achieved by minimising the negative log-likelihood in human preference data;
\begin{equation}
\begin{split}
    \mathcal{L} (r_\phi) =  -\mathbb{E}_{(x,y_w,y_l)\sim \mathcal{D}} &\big[ \log \sigma (r_\phi(x,y_w) \\
    & - r_\phi(x,y_l)) \big]
\end{split}
\end{equation}
Additionally, the Kullback-Leibler (KL) divergence between the outputs generated by $\pi_{\text{ref}}$ and the parameterised $\pi_{\theta}$ models serves as an additional reward signal, ensuring that the generated responses closely align with the reference model. Consequently, an optimal model $\pi_{\theta}$ is one that maximises;
\begin{equation}
\begin{split}
    \mathbb{E}_{(x \in \mathcal{D}, y \in \pi_{\theta})}[r_{\phi}(x,y)] 
    &- \beta \mathcal{D}_{\text{KL}}\bigl(\pi_{\theta}(y \mid x) \\
    &\quad\; || \pi_{\text{ref}}(y \mid x)\bigr)
\end{split}
\end{equation}

where $\beta$ is parameter typically $\in [0.1, 0.5]$.

\end{itemize}

\paragraph{Human-aware Loss Functions (HALOs):} 
\begin{definition}[HALOs]
 Let $x \in X$ and $y \in Y$ denote an input and output respectively. An $f: (x,y) \rightarrow \mathbb{R}$ is considered a human-aware loss function if it satisfies
\begin{equation}
\begin{split}
    f(x, y; \theta) &= t\Bigl(v_f \bigl(r_{\theta}(x, y) \\
    &- \mathbb{E}_{x'\sim Q',y'\sim Q'}[r_{\theta}(x', y')] \bigr)\Bigr)
\end{split}
\end{equation}
with a parameterised reward function $r_{\theta}$ such that $\forall (x_1,y_1),(x_2,y_2) \in X \times Y$, $r_{\theta}(x_1,y_1) > r_{\theta}(x_2,y_2) \Leftrightarrow (x_1,y_1) \succ_{r_{\theta}} (x_2,y_2)$, reference point distributions $Q_x(X')$ and $Q_y(Y'|X')$, a value function $v_f : \mathbb{R} \rightarrow \mathbb{R}$ that is monotonic non-decreasing and concave in $(0, \infty)$, and a negative affine function $t$.
\end{definition}

RLHF can present challenges due to inherent slowness and instability, especially in the case of highly varied outputs~\cite{zheng2024balancing}. Recently, there has been a shift towards using closed-form losses in RLHF to align LLMs with human preferences. These losses are predominantly HALOs that model human biases, as discussed in~\cite{tversky1992advances}, aiming to maximise the margin between preferred and dispreferred generated outputs. This approach offers a mathematical equivalence with RLHF, while effectively addressing inherent limitations.

\section{Data}
\label{appx:data}
Experiments are conducted on the benchmark \textit{L+M-24}, which integrates molecular and linguistic modalities in four key categories: biomedical; light and electricity; human interaction and organoleptics; and agriculture and industry~\cite{edwards2024l+}. The dataset was created by extracting SMILES (Simplified Molecular-Input Line-Entry System) strings and their chemical properties from databases. The properties were transformed into natural language using GPT-4 templates. The training and validation subsets contain around 127k and 34k language-molecule pairs, with only 10\% used for our experiments.

For evaluation, we randomly selected 3k unseen pairs from a separate dataset provided by the L+M-24 research group. The samples were drawn from 635k molecule-caption pairs included in the supplementary data for the Language + Molecules @ ACL2024 workshop~\footnote{\url{https://github.com/language-plus-molecules/LPM-24-Dataset}}.

\section{LLM Instruction for Decomposing Text into Atomic Context Units }
\label{appx:instructions_eval}

\begin{figure}[h!]
\centering
\begin{tcolorbox}[colback=gray!4!white,colframe=black!75!black,fontupper=\small]
  \#\#\# User:\\ Please breakdown the following text into independent facts: \\
 \{It impacts both atherosclerosis and cardiovascular disease. The molecule is a nutrient, thyroxine treatment, fat storage that impacts pancreatitis and metabolic syndrome.\}\\ 
 
 **\\
IMPORTANT: No words or explanation is needed. Only include truths that are factual.\\
You should NOT include any prior knowledge, and take the text at face value when extracting facts.\\
 **
            
  \#\#\# Assistant: \\
  - The molecule impacts atherosclerosis. 
- The molecule impacts cardiovascular disease. 
- The molecule is a nutrient. 
- The molecule is a thyroxine treatment. 
- The molecule impacts pancreatitis. 
- The molecule impacts metabolic syndrome. 
- The molecule is involved in fat storage. 
\end{tcolorbox}
\caption{Instruction for decomposing text into atomic content units conveying one piece of information. }
\label{nli_instruction}
\end{figure}

\section{Instructions for Molecule Language Translation}
\label{appx:instructions}

\begin{figure}[h!]
\centering
\begin{tcolorbox}[colback=gray!4!white,colframe=black!75!black,fontupper=\small]
  Below is an instruction that describes a task, paired with an input that provides further context.\\
  Write a response that appropriately completes the request. \\ \\
  \#\#\# Instruction: You are a researcher. You can come up captions based on your existing knowledge. \\ Captions are given against the following input. You should be as detailed as possible. \\ \\
  \#\#\# Input: Molecule: \{\textcolor{blue}{source molecule}\} \\
  In that molecule, could you formulate a caption about? \\ \\ \\ 
  \#\#\# Response:\{\textcolor{blue}{target caption}\}
\end{tcolorbox}

\caption*{Instruction for caption generation, i.e., $M \rightarrow L$}
\end{figure}

\begin{figure}[h!]
\centering
\begin{tcolorbox}[colback=gray!4!white,colframe=black!75!black,fontupper=\small]
  Below is an instruction that describes a task, paired with an input that provides further context.\\
  Write a response that appropriately completes the request. \\ \\
  \#\#\# Instruction: You are a researcher. You can come up molecule smile strings based on your existing knowledge. \\
  Molecule smile strings are given against the following input. You should be as detailed as possible.\\ \\
  \#\#\# Input: Caption: \{\textcolor{blue}{source caption}\} \\
  In that caption, could you generate a molecule smile string? \\ \\ \\ 
  \#\#\# Response: \{\textcolor{blue}{target molecule}\}
\end{tcolorbox}

\caption*{Instruction for molecule generation, i.e., $L \rightarrow M$}
\end{figure}

\FloatBarrier

\section{Baselines}
\label{appx:baselines}

\begin{itemize}[noitemsep,topsep=0pt,parsep=0pt,partopsep=0pt,leftmargin=*] 
\item \textit{TxtChem-T5}~\cite{christofidellis2023unifying} is a T5$_{XL}$ multitask model trained on linguistic and molecule modalities across multiple datasets, including CheBI-20, akin to L+M-24. 
\item \textit{Chem-LLM}~\cite{zhang2024chemllm}, an InternLM2-Base-7B model, is trained on large chemical knowledge databases using DPO, achieving GPT-4-level results. \item \textit{Meditron}~\cite{chen2023meditron}, a 7B model, is fine-tuned on the entire L+M-24 dataset.
\end{itemize}

\section{Evaluation Metrics}
\label{appx:metrics}
For performance evaluation, we employ established metrics from the literature~\cite{setsbenchmarking,edwards2022translation}. In translation from molecule to language, we assess using BLEU-2, BLEU-4, ROUGE-1, ROUGE-2, ROUGE-L, and METEOR metrics. For translation from molecule to language, evaluation metrics include BLEU, Levenshtein distance, fingerprint metrics (MACCS, RDK, and Morgan), Fréchet ChemNet Distance (FCD), and molecule validity metrics. The annotations in the result tables indicate whether higher or lower values indicate superior performance.

\section{Training Efficiency}
\label{appx:efficiency}

\begin{figure}[h!]
\centering
\includegraphics[width=1\columnwidth]{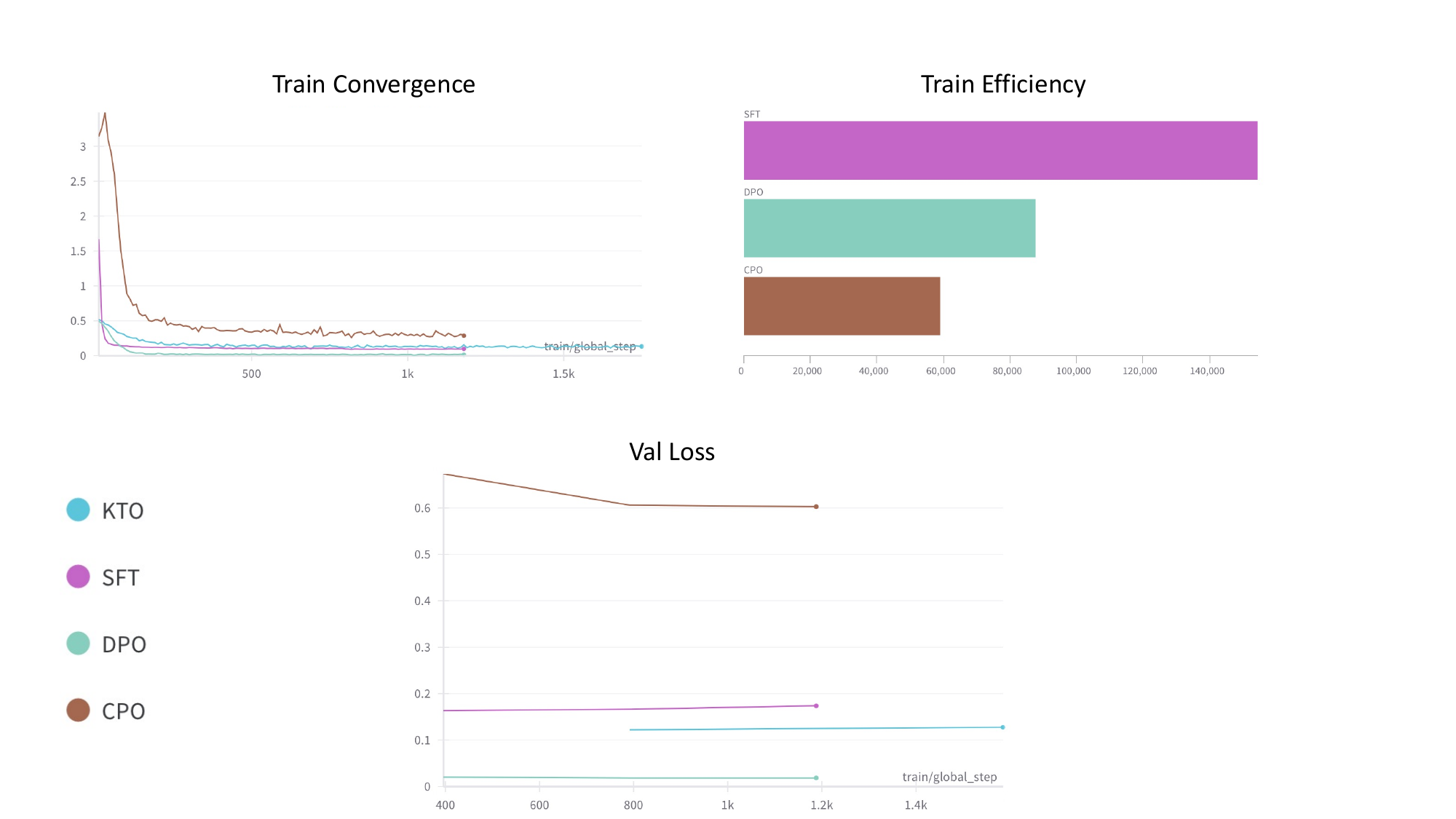}

\caption{Training efficiency across alignment fine-tuning methods }
\end{figure}

\section{Implementation Details}
\label{appx:implementation}

All implementations used Meditron~\cite{chen2023meditron} as the backbone model, known for its performance on L+M-24. For alignment fine-tuning experiments, we initialised Meditron crossmodals, trained for molecule generation~\footnote{Crossmodal initialisation was based on the most challenging task reported in \cite{edwards2024l+}.}. For the model merging experiments, we combined Meditron weights trained on MoCG tasks in a 1:18 ratio. This ratio aimed to preserve the balance of information between the linguistic and molecule modalities. All models were fine-tuned using QLoRA~\cite{dettmers2024qlora}.

For the atomic-level NLI evaluation method,  we instruct Meta-Llama-3-8B~\cite{touvron2023llama} to break down $(\text{reference, generated})$ pairs into a series of atomic premises and hypotheses. We then use DeBERTa~\footnote{https://huggingface.co/MoritzLaurer/DeBERTa-v3-large-mnli-fever-anli-ling-wanli} to measure hallucination and coverage by performing NLI across all the atomic premises and hypotheses.

\begin{figure}[h!]
    \centering
    \begin{tcolorbox}[colback=gray!4!white,colframe=black!75!black,fontupper=\small]
   
(\\
    load_in_4bit=True,\\
    bnb_4bit_use_double_quant=True,\\
    bnb_4bit_quant_type=nf64,\\
      bnb_4bit_compute_dtype=torch.bfloat16\\
)
\end{tcolorbox}

    \caption{Quantisation Configurations}
\end{figure}

\begin{figure}[h!]
    \centering
    \begin{tcolorbox}[colback=gray!4!white,colframe=black!75!black,fontupper=\small]
    
args = TrainingArguments(\\
    output_dir=save_path, \\ 
    overwrite_output_dir=True,\\
    load_best_model_at_end=True,\\
    num_train_epochs=3,       \\
    per_device_train_batch_size=1\\
    per_device_eval_batch_size=1\\
    gradient_accumulation_steps=64\\
    gradient_checkpointing=False\\
    optim="adamw_torch_fused",   \\            
    learning_rate=5e-5, \\              
    max_grad_norm=0.3,  \\        
    warmup_ratio=0.1, \\         
    lr_scheduler_type="cosine", \\            

)
\end{tcolorbox}

    \caption{Training configurations}
\end{figure}

\begin{figure}[h!]
    \centering
    \begin{tcolorbox}[colback=gray!4!white,colframe=black!75!black,fontupper=\small]

(\\
    lora_alpha=16,\\
    r = 64,\\
    lora_dropout=0.1,\\
    task_type="CAUSAL_LM",\\
    bias=False,\\
    target_modules= "all-linear" \\
)\\
\end{tcolorbox}

    \caption{LoRA Configurations}
\end{figure}

\FloatBarrier
\section{Examples of generated molecules and captions.}
\label{appx:examples_gen}

\begin{figure*}[h!] 
  \centering
  \includegraphics[width=1\textwidth]{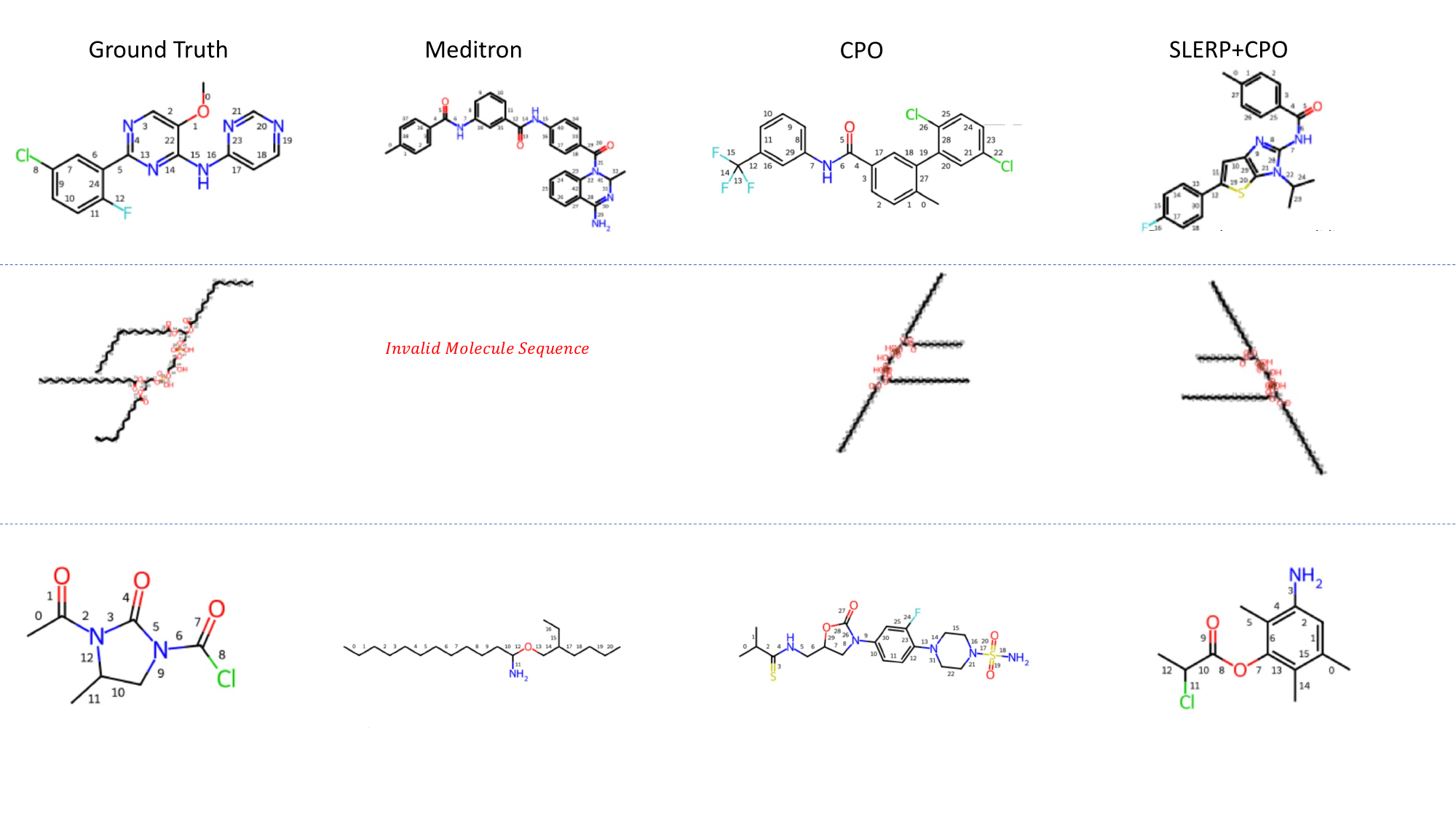} 
  \caption{Examples of molecules generated by our top-performing models compared to Meditron, the best benchmark model trained on the entire dataset.}
  \label{fig:gen_molecules}
 
\end{figure*}

\begin{figure*}[h!] 
  \centering
  \includegraphics[width=1\textwidth]{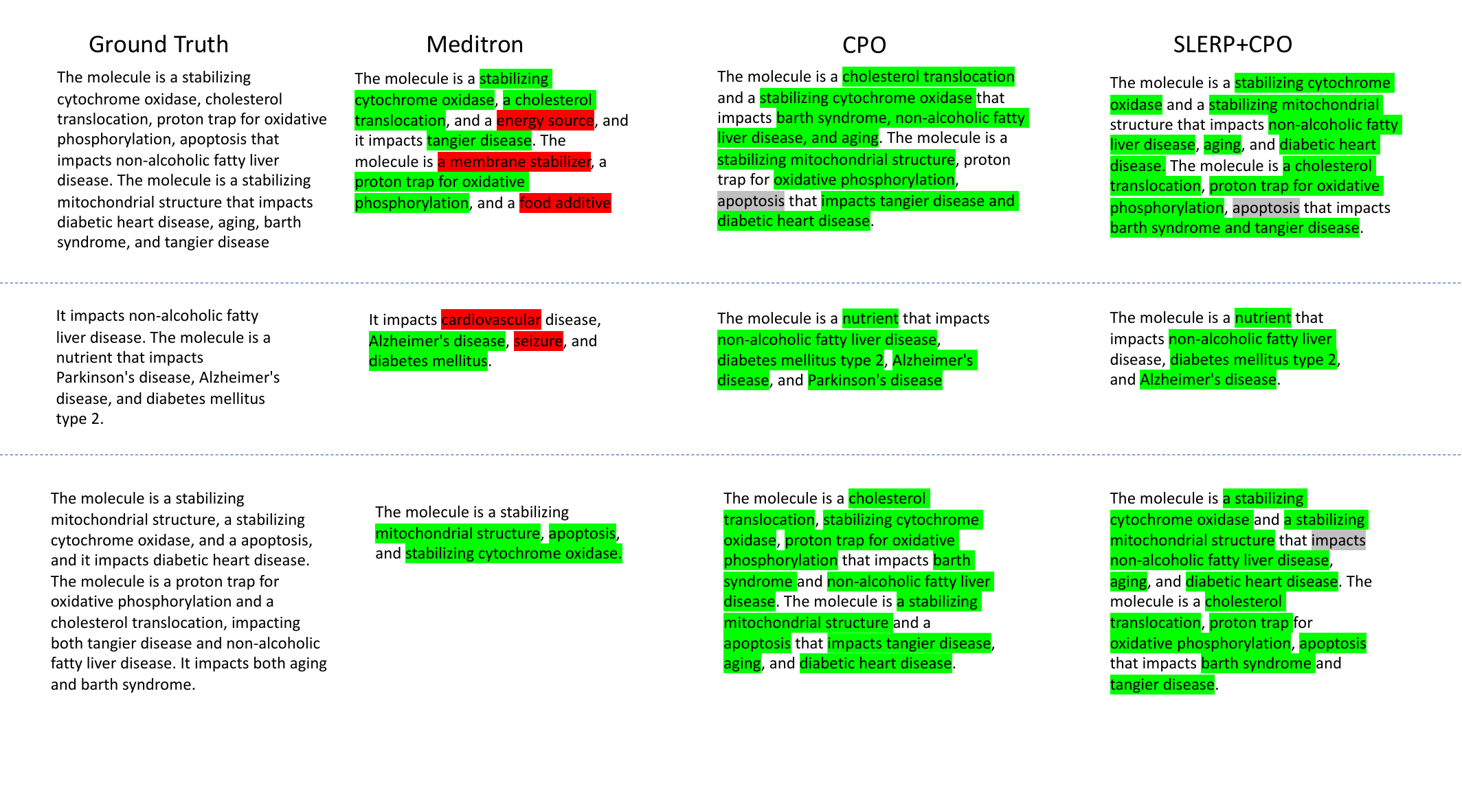} 
  \caption{Examples of captions generated by our top-performing models compared to Meditron, the best benchmark model trained on the entire dataset.}
   \label{fig:gen_caption} 
\end{figure*}

Fig.~\ref{fig:gen_molecules} and \ref{fig:gen_caption} illustrate examples of molecules and captions generated by our top-performing models compared to Meditron, respectively.

\section{Examples of Atomic-level Cross-NLI evaluation}
\label{appx:nli_cases}

Table~\ref{tbl:nli_examples} presents examples of assessing hallucination and coverage in generated captions using our atomic-level cross-NLI evaluation method.

\begin{table*}[b]
  \centering 
  \resizebox{\textwidth}{!}{%
  \begin{tabular}{p{5cm}p{5cm}p{5cm}p{5cm}cc }
    \toprule

    \textbf{Reference Text} & \textbf{Atomic Premises}  & \textbf{Generated Text} &  \textbf{Atomic Hypothesis}  &   \textbf{Hallucination}  &  \textbf{Coverage}      \\ \hline

    It impacts pancreatitis. The molecule is a fat storage and nutrient, belonging to the thyroxine treatment class of molecules, and impacts metabolic syndrome, atherosclerosis, and cardiovascular disease. & 
    
    \textcolor{Red}{- The molecule impacts pancreatitis.} \newline
     \textcolor{Red}{-The molecule is a fat storage molecule.}  \newline
     \textcolor{Green}{-The molecule is a nutrient.}\newline
    \textcolor{Red}{- The molecule belongs to the thyroxine treatment class of molecules.}\newline
    \textcolor{Red}{- The molecule impacts metabolic syndrome.} \newline
    \textcolor{Red}{- The molecule impacts atherosclerosis.} \newline
    \textcolor{Red}{- The molecule impacts cardiovascular disease.}

    & The molecule is a nutrient. &
   \textcolor{Green}{- The molecule is a nutrient.} &
   0.00 & 0.14
    \\ \hline
    The molecule is a energy storage and is floral. The molecule is a emulsifier, nutrient, surfactant, energy source, membrane stabilizer, and rose.
    
 & 
    
    \textcolor{Green}{- The molecule is a floral energy storage.} \newline
    \textcolor{Green}{- The molecule is an emulsifier.} \newline
    \textcolor{Green}{- The molecule is a nutrient.} \newline
    \textcolor{Green}{- The molecule is a surfactant.} \newline
    \textcolor{Green}{- The molecule is an energy source.} \newline
    \textcolor{Green}{- The molecule is a membrane stabilizer.} \newline
    \textcolor{Red}{- The molecule is rose.}

    & The molecule is a energy storage, a membrane stabilizer, and a energy source. The molecule is a surfactant, a emulsifier, and a nutrient. &
   \textcolor{Green}{- The molecule is an energy storage.} \newline
   \textcolor{Green}{- The molecule is a membrane stabilizer.} \newline 
    \textcolor{Green}{- The molecule is an energy source.} \newline
     \textcolor{Green}{- The molecule is a surfactant.} \newline
     \textcolor{Green}{- The molecule is an emulsifier.} \newline
      \textcolor{Green}{- The molecule is a nutrient.} \newline &
   0.00 & 0.75
    \\ \hline

    The molecule is a orexin receptor antagonist.
    
 & 
    
    \textcolor{Red}{- The molecule is an orexin receptor antagonist.} \newline

    & The molecule is a anti viral. &
   \textcolor{Red}{- The molecule is an anti-viral.} &
   0.75 & 0.00
    \\ \hline
    The molecule is a stabilizing cytochrome oxidase, apoptosis, stabilizing mitochondrial structure that impacts non-alcoholic fatty liver disease and tangier disease. The molecule is a cholesterol translocation and a proton trap for oxidative phosphorylation that impacts aging, barth syndrome, and diabetic heart disease.
    
 & 
    
    \textcolor{Red}{- The molecule is a cytochrome oxidase.} \newline
    \textcolor{Green}{- The molecule is a stabilizer of apoptosis.} \newline
    \textcolor{Green}{- The molecule is a stabilizer of mitochondrial structure.} \newline
    \textcolor{Green}{- The molecule impacts non-alcoholic fatty liver disease.} \newline
    \textcolor{Green}{- The molecule impacts Tangier disease.} \newline
    \textcolor{Green}{- The molecule is a cholesterol translocation.} \newline
    \textcolor{Red}{- The molecule is a proton trap.} \newline
    \textcolor{Green}{- The molecule impacts oxidative phosphorylation.} \newline
    \textcolor{Green}{- The molecule impacts aging.} \newline
    \textcolor{Green}{- The molecule impacts Barth syndrome.} \newline
    \textcolor{Green}{- The molecule impacts diabetic heart disease.} 
    & The molecule is a cholesterol translocation, a apoptosis, and a stabilizing cytochrome oxidase, and it impacts tangier disease. The molecule is a stabilizing mitochondrial structure and a proton trap for oxidative phosphorylation that impacts barth syndrome, aging, and non-alcoholic fatty liver disease. It impacts diabetic heart disease. &
   \textcolor{Green}{- The molecule is a cholesterol translocation.} \newline
   \textcolor{Green}{- The molecule is involved in apoptosis.} \newline
   \textcolor{Green}{- The molecule is a stabilizing cytochrome oxidase.} \newline
   \textcolor{Green}{- The molecule impacts Tangier disease.} \newline
   \textcolor{Green}{- The molecule is a stabilizing mitochondrial structure.} \newline
   \textcolor{Green}{- The molecule is a proton trap for oxidative phosphorylation.} \newline
   \textcolor{Green}{- The molecule impacts Barth syndrome.} \newline
   \textcolor{Green}{- The molecule impacts aging.} \newline
   \textcolor{Green}{- The molecule impacts non-alcoholic fatty liver disease.} \newline
   \textcolor{Green}{- The molecule impacts diabetic heart disease.} 
      
   &
   0.00 & 0.91
    \\ \hline
    The molecule is a anti microbial member of the anti fungal class. & 
    \textcolor{Green}{- The molecule is anti-microbial.} \newline
    \textcolor{Red}{- The molecule is a member of the anti-fungal class.} \newline
    & 
    It belongs to the anti viral class of molecules. The molecule is both a hepatitis c treatment and a hcv inhibitor.
    & 
    \textcolor{Green}{- The molecule belongs to the anti-viral class of molecules.} \newline
    \textcolor{Green}{- The molecule is a hepatitis C treatment.} \newline
    \textcolor{Green}{- The molecule is an HCV inhibitor.} \newline 
    & 0.02& 0.10 \\

  \bottomrule
  \end{tabular}
 }
  \caption{ Cases showcasing insights captured by our atomic-level cross-NLI in assessing the level of hallucination and coverage in generated captions. Red highlights indicate missing information in atomic premises or invalid information in atomic hypotheses. Hallucination refers to the introduction of information absent from the reference, while coverage assesses the recall of atomic units (refer to $\S$~\ref{sec:atomic_nli}). }
  \label{tbl:nli_examples}
\end{table*}

\end{document}